\documentclass[a4paper,fleqn]{cas-sc}

\usepackage[authoryear,longnamesfirst]{natbib}
\usepackage{rotating}
\usepackage{tabularx}
\usepackage{makecell} 
\usepackage{booktabs}
\usepackage{array}

\def\tsc#1{\csdef{#1}{\textsc{\lowercase{#1}}\xspace}}
\tsc{WGM}
\tsc{QE}
\tsc{EP}
\tsc{PMS}
\tsc{BEC}
\tsc{DE}
\begin{document}
\let\WriteBookmarks\relax
\def\floatpagepagefraction{1}
\def\textpagefraction{.001}

\shorttitle{Leveraging social media news}

\shortauthors{CV Radhakrishnan et~al.}

\title [mode = title]{DeepMEL: A Multi-Agent Collaboration Framework for Multimodal Entity Linking}


\author[1]{Fang Wang}[bioid=1]
\fnmark[1]
\ead{fangwang@stu.pku.edu.cn}

\author[2]{Tianwei Yan}[style=chinese]
% email: augusyan@hotmail.com
\author[3]{Zonghao Yang}[style=chinese]
% email: yangzonghao1024@163.com
\author[4]{Minghao Hu}[style=chinese]
% email: huminghao16@gmail.com
\author[4]{Jun Zhang}[style=chinese]
% email: mcgrady150318@163.com
\author[4]{Zhunchen Luo}[style=chinese]
% email: zhunchenluo@gmail.com
\author[4]{Xiaoying Bai}[style=chinese]
% email: baixy@aibd.ac.cn

\affiliation[1]{organization={School of Computer Science, Peking University},
    addressline={No.5 Yiheyuan Road}, 
    city={Beijing},
    postcode={100871}, 
    country={China}}

\affiliation[2]{organization={School of Information Science and Engineering, Chongqing Jiaotong University},
    addressline={No.66 Xuefu Road}, 
    city={Chongqing},
    postcode={400074}, 
    country={China}}

\affiliation[3]{organization={China Research and Development Academy of Machinery Equipment},
    addressline={No. 10 Courtyard Road}, 
    city={Beijing},
    postcode={100089}, 
    country={China}}

\affiliation[4]{organization={Center of Information Research, Academy of Military Science},
    addressline={No.26 Fucheng Road}, 
    city={Beijing},
    postcode={100142}, 
    country={China}}

\cortext[cor1]{Corresponding author}

\fntext[fn1]{This is the first author footnote.}

\begin{abstract}
%Multimodal Entity Linking (MEL) aims to fuse text and visual information and accurately associate real-world mentions with corresponding entities in a multimodal knowledge graph (MMKG). It is a crucial technology for enhancing the performance of downstream tasks in natural language processing (NLP). However, existing methods generally face challenges such as incomplete contextual information, insufficient granularity in cross-modal feature fusion, and difficulties in jointly optimizing large language models (LLMs) and large visual models (LVMs). 
Multimodal Entity Linking (MEL) aims to associate textual and visual mentions with entities in a multimodal knowledge graph. Despite its importance, current methods face challenges such as incomplete contextual information, coarse cross-modal fusion, and the difficulty of jointly large language models (LLMs) and large visual models (LVMs). To address these issues, we propose DeepMEL, a novel framework based on multi-agent collaborative reasoning, which achieves efficient alignment and disambiguation of textual and visual modalities through a role-specialized division strategy. DeepMEL integrates four specialized agents, namely Modal-Fuser, Candidate-Adapter, Entity-Clozer and Role-Orchestrator, to complete end-to-end cross-modal linking through specialized roles and dynamic coordination. DeepMEL adopts a dual-modal alignment path, and combines the fine-grained text semantics generated by the LLM with the structured image representation extracted by the LVM, significantly narrowing the modal gap. We design an adaptive iteration strategy, combines tool-based retrieval and semantic reasoning capabilities to dynamically optimize the candidate set and balance recall and precision. DeepMEL also unifies MEL tasks into a structured cloze prompt to reduce parsing complexity and enhance semantic comprehension. Extensive experiments on five public benchmark datasets demonstrate that DeepMEL achieves state-of-the-art performance, improving ACC by 1\%-57\%. Ablation studies verify the effectiveness of all modules.
\end{abstract}

% Keywords

\maketitle

%%--------------------Introduction--------------------
\section{Introduction}
Entity linking is a fundamental task in knowledge graph (KG) construction~\cite{Construction}, aiming to link mentions to their corresponding entities in a target knowledge base (KB). It is widely applied in downstream natural language processing (NLP) tasks, such as Question \& Answering Systems~\cite{qa3} and intelligent recommendation systems~\cite{recommand-system-2}. Recently, the explosive growth of multimodal data on the Internet has raised challenges, as the quality of online information is often inconsistent, many mentions are ambiguous, and contextual information is frequently incomplete. Under such conditions, relying solely on a single modality (such as pure text) is often insufficient to accurately resolve reference ambiguity~\cite{gan2021multimodal}. Integrating textual and visual modalities can significantly improve the precision and efficiency of disambiguation~\cite{gella2017disambiguating}. Consequently, multimodal entity linking, which involves combining textual and visual information to link real-world mentions to corresponding entities in a multimodal knowledge graph (MMKG), has become a critical research task.

\begin{figure}[htpb]
    \centering
    \includegraphics[scale=.65]{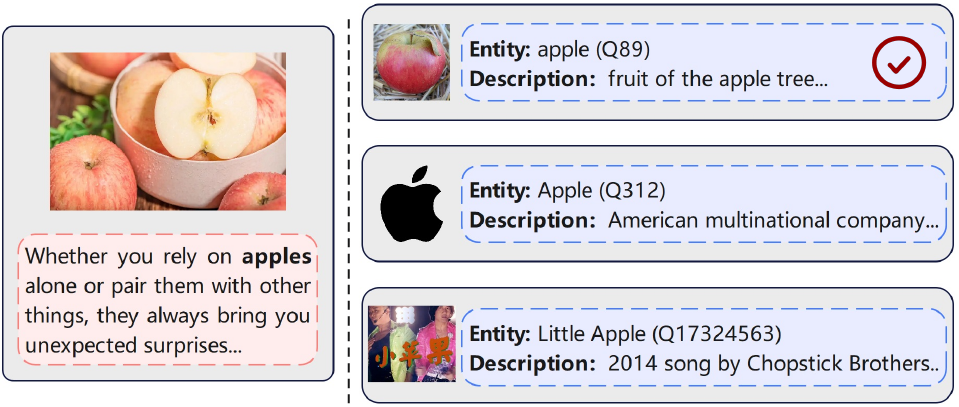}
    \caption{An example of Multimodal Entity Linking. Left: mention image and mention text. Right: the candidates with its image and description.}
    \label{fig:intro-sample}
\end{figure}

For example, as shown in Figure~\ref{fig:intro-sample}, the mention of "Apple" may be difficult to disambiguate, as it could refer to various entities, such as Apple Inc. or the apple (fruit). However, by considering both textual and visual information, it becomes easier and clearer to accurately link the mention of "Apple" to the entity "apple (fruit of the apple tree)."

Currently, multimodal entity linking models are primarily based on deep learning frameworks, utilizing cross-attention mechanisms~\cite{cross-attention} and visual feature encoding techniques~\cite{visual-encoding2} to achieve the fusion of textual mentions and visual information. 

With the rapid development of Large Language Models (LLMs)~\cite{LLM3} and Large Visual Models (LVMs)~\cite{LVM3}, methods based on large models have gradually become a research hotspot. However, existing methods still face the following key challenges:

(1) Long-distance dependency and fragmentation of contextual information. As shown in Figure~\ref{fig:intro-sample}, the mention's context often contains redundancy and discontinuity, while the model input is typically limited to truncated fragments of the document. This results in the key semantic clues and discriminative evidence required for entity linking being scattered in different fragments and difficult to be effectively captured.

(2) Insufficient granularity of cross-modal feature fusion. Although text and visual modalities theoretically possess complementary advantages, existing studies typically adopt a fusion-then-matching strategy. This involves separately extracting multimodal representations for mentions and entities, and then performing entity linking through similarity computation. However, these methods fuse text and image features before the matching process, which prevents them from fully exploiting the fine-grained alignment relationships between mentions and entities.

(3) Limitations of large model collaborative optimization. Although LLMs excel in text understanding tasks, their visual perception ability is limited by the purely text-based training paradigm. On the other hand, LVMs, although possessing fine-grained visual understanding, struggle with complex semantic reasoning. Designing an efficient collaboration mechanism that allows LLMs to dynamically integrate the visual capabilities of LVMs in knowledge reasoning remains an open problem.

In order to solve the above problems, we proposed a novel multimodal entity linking framework, DeepMEL (A Multi-Agent Collaboration Framework for Multimodal Entity Linking). This framework is based on a Multi-Agent Collaborative Reasoning mechanism and employs a Role-Specialized Division strategy to achieve cross-modal feature fusion and entity disambiguation. DeepMEL leverages the collaborative interaction between LLMs and LVMs to achieve end-to-end multi-modal entity linking, significantly improving entity recognition accuracy in heterogeneous data environments. Specifically, DeepMEL consists of four roles: Modal-Fuser, Candidate-Adapter, Entity-Clozer, and Role-Orchestrator. The Modal-Fuser leverages the LLMs’ summarization ability to generate concise, information-rich context and uses the LVMs’ visual question-answering function to extract structured semantic descriptions of image entities, enabling efficient alignment from visual to text modality. The Candidate-Adapter dynamically filters candidates based on embedding retrieval tools and progressively optimizes the Top-K candidate set through an Adaptive Iteration strategy, ensuring a balance between high recall and high precision. The Entity-Clozer introduces a cloze-style prompt to enhance the LLM’s ability to perform MEL. The Role-Orchestrator, as the central scheduling module, coordinates task distribution and model updates among the agents, ensuring the dynamic adaptability of the framework through the Role-specialized division strategy.

Overall, the contributions of our work can be summarized as follows:

1. We propose DeepMEL, a multi-agent framework based on Role-Specialized, which fuses visual and context information through collaboration between LLMs and LVMs. To the best of our knowledge, this is the first multi-agent solution applied to MEL, thereby expanding the task’s modeling paradigm. 

2.We propose a modality conversion alignment strategy for cross-modal representation alignment through two paths: using a Context Summary Prompt with LLMs to generate fine-grained descriptions, and a Visual Q\&A Prompt with LVMs to extract structured image representations.

3. We propose an Adaptive Iteration strategy that dynamically optimizes candidates by combining tool-based retrieval with LLM-based semantic reasoning.

4. We propose a unified prompt that reformulates MEL into a structured cloze-style task, simplifying input parsing and improving LLMs' semantic understanding of multimodal disambiguation.

5. We empirically evaluate DeepMEL on five public benchmark datasets, where it consistently outperforms existing methods with accuracy improvements ranging from 1\% to 57\%. In addition, ablation studies confirm the effectiveness of each module, underscoring the importance of our multi-agent design.

%%--------------------Introduction--------------------
\section{Related Work}
\subsection{Textual Entity Linking}
Entity Linking (EL), as a fundamental task in Natural Language Processing (NLP), has long been a focus of extensive research in the academic community. 

Recent methods for Entity Linking (EL) all employ neural network-based architectures. A typical framework generally comprises the following key components: First, context-aware representations of mentions and entities are generated using either pre-trained or sequence-based text encoders (e.g., LSTM~\cite{LSTM} or BERT~\cite{BERT}). Subsequently, semantic similarity is computed via vector-space metrics such as dot product~\cite{KolitsasGH18,wu2019scalable} or cosine similarity~\cite{SunLTYJW15}. Finally, some studies incorporate a feed-forward network (FFN) combined with a softmax normalization layer~\cite{Entity-aware-ELMo} to produce a probabilistic prediction of the distribution of entities.

To systematically evaluate the performance of EL models, researchers have constructed various benchmark datasets, including high-quality human-annotated datasets AIDA~\cite{hoffart-etal-2011-robust}, large-scale automatically labeled datasets CWEB~\cite{guo2018robust}, and evaluation sets designed for zero-shot scenarios Zeshel~\cite{logeswaran-etal-2019-zero}. However, existing studies indicate that the performance of mainstream EL models on traditional benchmarks (e.g., AIDA-test, MSNBC~\cite{cucerzan-2007-large}, and AQUAINT~\cite{milne2008learning}) is approaching theoretical upper bounds, suggesting that these evaluation sets may have reached saturation. In particular, with the advancement of large-scale pre-trained language models, state-of-the-art approaches have achieved F1 scores exceeding 90\% on the AIDA dataset, further confirming the emergence of performance bottlenecks~\cite{de2020autoregressive}. To overcome this limitation, recent research has shifted toward more challenging directions, including zero-shot entity linking~\cite{logeswaran-etal-2019-zero,wu2019scalable}, global consistency modeling, NIL entity prediction~\cite{rao2013entity}, and end-to-end recognition and linking of emerging entities~\cite{broscheit2020investigating}. These novel paradigms are expected to drive EL technology toward deeper advancements, pushing beyond the current performance limits of existing methods. 

Although text-based methods~\cite{feher2023few, leonard2024ufel, li2025collective} have achieved significant progress, they are inherently limited in handling multimodal tasks. Given the increasing prominence of multimodal data in recent years, there is a pressing need for novel approaches capable of addressing MEL challenges.

\subsection{Multimodal Entity Linking}
In recent years, Multimodal Entity Linking (MEL) has emerged as a critical research direction in the field of Natural Language Processing (NLP). The core challenge of this task lies in accurately disambiguating and mapping ambiguous mentions from unstructured data to their corresponding entities in a Knowledge Graph (or Knowledge Base) by effectively fusing textual and visual modalities.  

The pioneering study by Moon et al.~\cite{moon-etal-2018-multimodal-named}first demonstrated the effectiveness of visual information in disambiguating short-text mentions on social media, proposing a zero-shot learning framework that integrates textual, visual, and lexical features. However, due to GDPR compliance requirements, their constructed dataset was not made publicly available. Subsequently, Adjali et al.~\cite{adjali-etal-2020-building}introduced an automated MEL dataset construction method based on Twitter data, but it exhibited notable limitations, including a restricted data scale, insufficient coverage of entity types, and low mention ambiguity levels. Zhou et al.~\cite{ZhouWLXW21} released three MEL datasets based on platforms such as Weibo and Richpedia and proposed a MEL dataset construction method. Zhang et al.~\cite{zhang2021attention}constructed a Weibo multimodal dataset in Chinese that exclusively focuses on person entity recognition, while Gan et al.~\cite{gan2021multimodal} formulated the alignment between textual and visual mentions as a bipartite graph matching problem and released a dataset built upon movie reviews. However, due to the limited diversity of entity types in these datasets, the trained models suffer from significantly constrained domain adaptability and generalization capability. Wang et al.~\cite{wang2022wikidiverse} employed a multimodal co-attention mechanism to hierarchically extract jointly attended textual and visual features. To address the limitations of restricted contextual topics and entity types, they attempted to construct WikiDiverse, a dataset encompassing diverse multimodal tasks, contextual themes, and multi-entity types, but there are still key problems such as insufficient data scale and imperfect multimodal task system. In order to model visual and textual information at the knowledge level, Zhang et al.~\cite{ZhangH22} acquired knowledge through meta-learning, unified multimodal representation through transfer learning, and proposed an Interactive Multimodal Learning Network (IMN) to make full use of multimodal information in terms of mentions and knowledge. Xing et al.~\cite{XingZWLZD23} introduced a novel framework termed Dynamic Relation Interactive Network (DRIN), which effectively captures fine-grained and dynamically adaptive alignment relationships between entities and mentions. Luo et al.~\cite{LuoXWZXC23} proposed a novel Multi-GraIned Multimodal InteraCtion Network (MIMIC) framework, which incorporates three key modules: Text-based Global-Local interaction Unit (TGLU), Vision-based DuaL interaction Unit (VDLU) and Cross-Modal Fusion-based interaction Unit (CMFU). This architecture is designed to comprehensively model both abbreviated textual context and implicit visual cues, enabling more robust multimodal representation learning. Shi et al.~\cite{LongZMZZ24} introduced a novel approach called GELR, which incorporates a knowledge retriever to enhance the visual information of the entity by leveraging external sources. Song et al.~\cite{SongZ00LMW24} formulated multimodal entity linking as a neural text matching problem, and implemented queries with multimodal data and addresses semantic gaps using cross-modal enhancers between text and image information.

Most of these works~\cite{zhao2025me3a, huang2025knowledge, zhang2025graph, ahmad2025benchmark} fuse the text and image on both the mention and entity side, and then use various techniques to match their information. Although these research studies have shown that visual information is beneficial to the performance of entity linking to some extent, this means that they implicitly model the alignment relation between the text and image of the mention and entity. 

\subsection{Agents-based Tasks}
Agents make decisions and take actions in dynamic, real-time environments to achieve some specific objectives. Large Language Models (LLMs) and Vision Language Models (LVMs), pre-trained on massive corpora, have demonstrated remarkable performance across various tasks, enabling their use in simulating real-world agent behaviors~\cite{ParkOCMLB23,XiaoZYLHYC23,huang2025papereval}. In particular, LLMs have proven highly effective in task planning~\cite{MindAgent}, encode substantial world knowledge~\cite{Kola}, and exhibit impressive logical reasoning capabilities~\cite{xiong2025afr}. For summarization tasks, LLMs can condense lengthy texts into concise summaries~\cite{IndicGenBench}. Although LLMs excel in text comprehension tasks, their visual perception remains constrained by their text-only training paradigm. To mitigate this limitation, there is growing interest in extending traditional LLMs into LVMs. Similarly, LVMs, pre-trained on vast amounts of visual and linguistic data, demonstrate strong capabilities in vision-centric tasks such as image captioning~\cite{blip2,OFA} and visual Question \& Answering~\cite{InstructBLIP,zhao2025towards}. However, while LVMs possess fine-grained visual understanding, they often exhibit deficiencies in complex semantic reasoning tasks.

%%--------------------Preliminary--------------------
\section{Preliminary}
\textbf{Mathematical notation. }Let \textit{D} be a single document from the document collection $\mathcal{D}$. Building upon the foundation laid by previous work on entity linking, we assume that relevant information pertaining to target mentions in document \textit{D} has been obtained through a named entity recognizer. The mention and its context are denoted as $M_i = (m_{w_i}, m_{s_i}, m_{v_i})$, where $m_{w_i}, m_{s_i}, m_{v_i}$ indicate the words of the mention, the summary of the mention, and the corresponding image, respectively. The target multimodal knowledge graph $\mathcal{KG}$ is constructed by a set of entities $\mathcal{E} = {\{E_j\}}_{j=i}^N$, and each entity is denoted as $E_j = (e_{n_j}, e_{v_j}, e_{d_j}, e_{a_j})$, where the elements of $E_j$ represent the entity name, the entity images, the entity description, and the entity attribute, respectively. The candidate set $\mathcal{C}$ is a subset of $\mathcal{E}$, where $\mathcal{C} \subseteq \mathcal{E}$. The surface form or semantics of the candidates are closer to mention.

\noindent \textbf{Task definition.} Given a mention $M_i$ from document \textit{D}, the task of multimodal entity linking targets to retrieve the candidate set $\mathcal{C}$ from entity set $\mathcal{E}$ of knowledge graph $\mathcal{KG}$, and links the ground truth entity $E_j$ from candidate set. Therefore, the multimodal entity linking can be defined as follows: 

\begin{equation}
    \Psi= \{(M_i, E_j) | M_i \in \textit{D}, E_j \in \mathcal{E}, M_i \leftrightarrow E_j\},
    \label{eq:1}
\end{equation}

where $M_i \leftrightarrow E_j$ represents target mention $M_i$ and the candidate entity $E_j$ are equivalent, i.e., $M_i$ and $E_j$ refer to the same real-world object.

%%--------------------Method--------------------
\section{Method}
\begin{sidewaysfigure}
    \centering
    \includegraphics[width=\textheight]{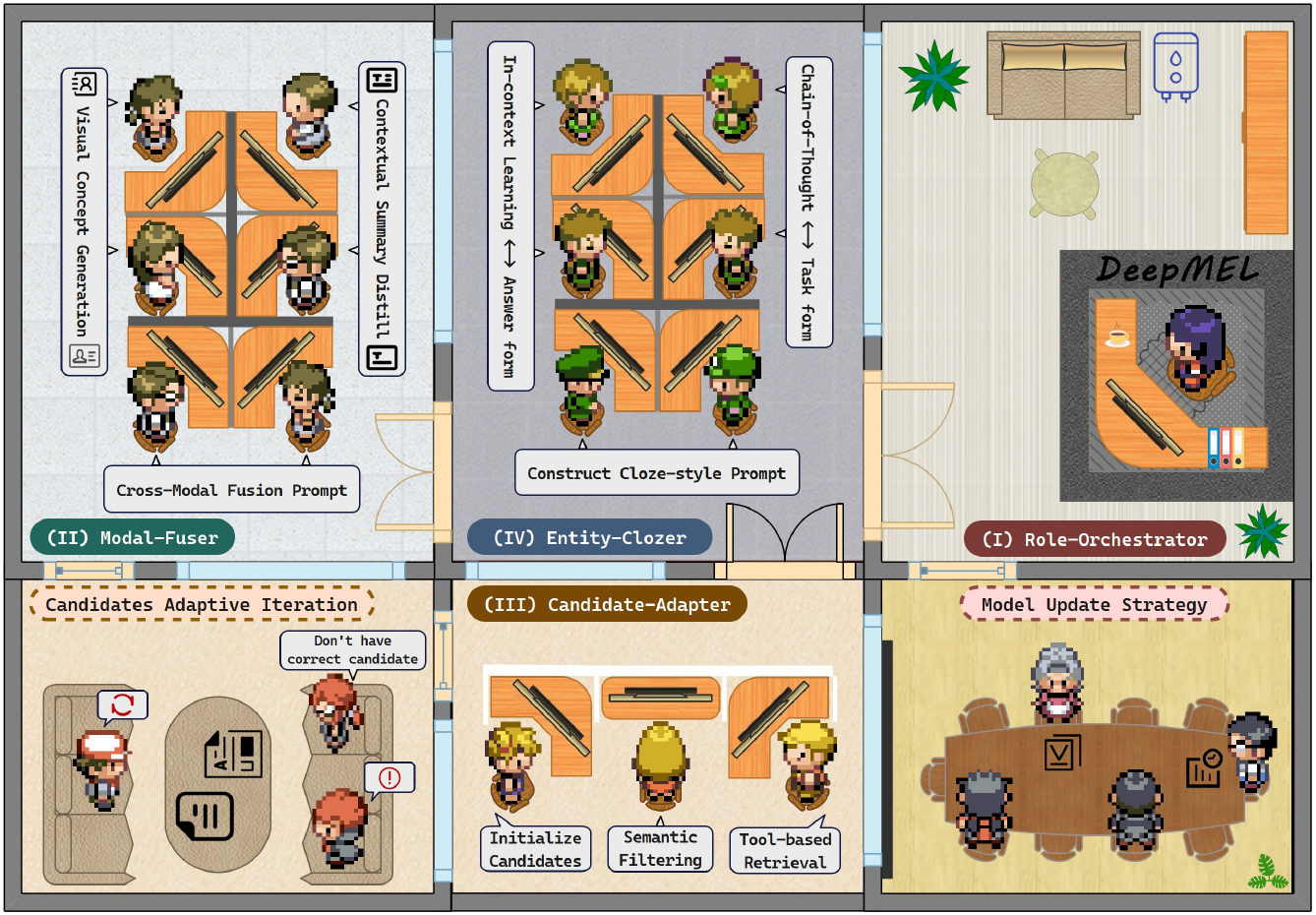}
    \caption{Overview of the DeepMEL. The Role-Orchestrator consists of Role-Specialized Division strategy and Model Update strategy. The Candidate-Adapter consists of dynamic candidate generation and Adaptive Iteration Strategy.}
    \label{fig:overview}
\end{sidewaysfigure}

In this section, we introduce DeepMEL, which aims to fuse multi-modality to improve the performance of multimodal entity linking. An overview of the framework is shown in Figure~\ref{fig:overview}.

\subsection{Overview}

Existing research on multimodal entity linking typically adopts a strategy of entity matching after modality fusion, where the multimodal representations of mentions and candidate entities are separately extracted, and entity linking is accomplished through similarity computation. However, since these approaches fuse textual and visual features before matching, they fail to fully leverage fine-grained alignment between mentions and candidates. Our framework employs a role-specialized multi-agent collaboration strategy to integrate visual and contextual information for MEL tasks. It achieves efficient alignment from the visual modality to the text modality through a modality conversion alignment strategy. In addition, the adaptive iteration strategy dynamically enhances the recall and precision of the candidate set. Finally, by transforming the task into a Cloze-style prompt, we improve the agent's comprehension and reasoning capabilities. In Section~\ref{sec:Role-Orchestrator}, we elaborate on how to coordinate multiple agents and subsequently update other LLMs and LVMs to enhance the robustness of MEL. In Section~\ref{sec:Modal-Fuser}, we analyze how to efficiently align visual and textual modalities. In Section~\ref{sec:Candidate-Adapter}, we discuss dynamically obtaining high-quality candidate sets. Finally, in Section~\ref{sec:Entity-Clozer}, we reformulate the entity disambiguation task as a cloze-style task to improve MEL performance.

\subsection{Role-Orchestrator: Multi-Agent Task Coordination}
\label{sec:Role-Orchestrator}

Based on the Role-Specialized Division strategy, our framework constructs a hierarchical multi-agent collaborative system that achieves optimized execution of multimodal entity linking through dynamic task allocation and coordination mechanisms. Specifically, the Role-Orchestrator employs the following core mechanisms.

\subsubsection{Specialized Task Decoupling}
% \textit{\textbf{Specialized Task Decoupling. } }
To handle the complexity of the MEL task, we decouple it into three specialized and interdependent sub-modules: modality fusion (Modal-Fuser), candidate generation (Candidate-Adapter), and entity disambiguation (Entity-Clozer). Each sub-module is implemented as a dedicated Agent optimized for its respective function. Specifically, Modal-Fuser aggregates and aligns multimodal information using LVMs; Candidate-Adapter retrieves and ranks candidate entities from large-scale knowledge sources guided by prompt-augmented LLMs; and Entity-Clozer performs fine-grained disambiguation based on context-aware reasoning and entity scoring. Importantly, the final decision is made by the Entity-Clozer, which integrates the fused representations and candidate information to select the most appropriate entity with high confidence. 

\subsubsection{Model Update Strategy}
%   \textit{\textbf{Model Update Strategy. }}
We design a Performance-Adaptive Update strategy that dynamically adjusts the model version or triggers model replacement for each agent based on its performance on the validation set. Specifically, each agent maintains a local performance tracker that monitors key metrics (e.g., task-specific accuracy) over a sliding evaluation window. If an agent's performance drops below a predefined degradation threshold, the system triggers a version update.

To ensure robustness and prevent instability during evaluation, all update decisions are governed by a strict rule-based supervision framework. This includes constraints such as:
(1) updates are only allowed within predefined evaluation intervals and (2) replacement candidates must pass prior validation under controlled settings.

\subsection{Modal-Fuser: Cross-Modal Alignment \& Fusion}
\label{sec:Modal-Fuser}
This module innovatively proposes the \textit{Dual-Path Representation Learning Strategy} to achieve precise alignment of visual-textual semantic space through a structured modality conversion strategy. The key components are as follows.

\subsubsection{Contextual Summary Distill}
% \textit{\textbf{Contextual Summary Distill. }}
Construct a \textit{Context Summary Prompt} is a specially designed input structure for large language models (LLMs), aimed at enabling fine-grained parsing of contextual semantics and efficient information distillation. The prompt is composed of three main components: the Task Understanding Module, the Demonstration Module, and the Format Specification Module. These components work in tandem to enhance the model’s capacity for semantic understanding and reasoning in complex contexts.

\begin{itemize}
    \item \textbf{Task Understanding Module.} This module incorporates task-oriented chain-of-thought (CoT) to explicitly define the LLM’s objectives and operational steps. It guides the LLM to deeply comprehend the input and summary relevant semantics.
    \item \textbf{Examples Module.} Leveraging an automated web crawler built on the Wikidata, this module selects 32 high-quality examples with comprehensive semantic coverage and balanced domain distribution. These examples, embedded within the prompt as contextual demonstrations, span various domains such as sports, politics, and entertainment. They effectively activate the model’s domain-specific prior knowledge and guide it to perform disambiguation, semantic alignment, and cross-context reasoning.
    \item \textbf{Format Specification Module.} This module enforces a standardized structure for both input and output, including summary formatting, paragraph boundaries, and field labeling. Such structural constraints enhance the consistency, controllability, and interpretability of the model’s outputs.
\end{itemize}

In summary, to achieve effective vision-language alignment and fusion, we first employ the Context Summary Prompt to distill the context surrounding each mention into a concise sentence containing key information. Then, using  Multi-Granularity Visual Question Answering Prompt, we generate textual descriptions for entities in the image, thereby transforming the visual entity linking task into a textual entity linking problem. This strategy effectively mitigates the issue of feature length mismatch caused by separate encoding of visual and textual modalities, and facilitates deep cross-modal integration.

\subsubsection{Visual Concept Generation}
% \textit{\textbf{Visual Concept Generation. }}
To enable accurate extraction of textual representations for key entities in images, we propose a Multi-Granularity Visual Question Answering prompt. This framework is designed to guide LVMs in a fine-grained, task-driven manner. The prompt consists of the following three core modules.

\begin{itemize}
    \item \textbf{Entity Specification Module.} Utilizes chain-of-thought reasoning to explicitly identify entities within the image, helping the model focus on relevant regions and enhancing the accuracy of entity recognition.
    \item \textbf{Granularity Control Module.} Refines the level of detail required for information extraction, ranging from high-level descriptions (e.g., identity, category) to fine-grained attributes (e.g., clothing color, facial expression, relative position).
    \item \textbf{Description Generate Module.} Constructs structured natural language questions that guide the model to generate semantic descriptions for the specified entities in the image. For example, given the entity “apple,” the generated question might be: “An apple is a type of fruit that can be red or green in color.”
\end{itemize}

In summary, the Context Summary Prompt establishes a systematic semantic distillation framework, empowering LLMs with stronger task adaptability in long-text comprehension, information selection, and structured output generation.

\subsection{Candidate-Adapter: Dynamic Candidate Generation}
\label{sec:Candidate-Adapter}

Candidate generation helps narrow down the search scope, thereby improving the efficiency of subsequent disambiguation. Searching directly across the entire knowledge Graph (knowledge Base) would incur substantial computational costs. However, by first generating candidates, only sorting and selecting from these candidates is necessary, which can significantly save computational resources.

\subsubsection{Tool-based Candidate Search} 
% \textbf{Tool-based Candidate Search.} 
We utilize the online version of Wikidata as the target knowledge graph and search for the target mention, retrieving the top-200 candidates related to it. These candidates include their names, Qid, descriptions, and other attributes. To further refine the candidate set and exclude irrelevant terms, we propose a filtering mechanism based on the similarity between the understanding of the mention and the descriptions of the candidates. Using BLEU~\cite{bleu} and BERTScore~\cite{sentence-bert} as similarity metrics, we calculate the similarity between the mention's understanding and the candidate descriptions, selecting the top-5 candidates with the highest similarity scores.

\subsubsection{Candidates Adaptive Iteration}
% \textbf{Candidates Adaptive Iteration. }
After obtaining the initial candidate set, the LLM is assigned the role of a Judgment Agent, responsible for evaluating whether the current candidate set contains the correct entity corresponding to the mention in the context. The judgment is based not only on the semantics of the mention itself but also on the contextual alignment between the mention and the candidates.

As illustrated in Figure~\ref{fig:overview}, if the candidate set includes the correct entity, the process proceeds to the cloze test stage for fine-grained verification. Otherwise, the current candidate set, along with the failure signals indicating misalignment with the mention, is fed back to the Modal-Fuser Agent. 

The Modal-Fuser Agent integrates information into the prompt, which includes:

\begin{itemize}
    \item A concise summary of the mention's contextual information;
    \item A semantic conflict analysis between the mention and the candidate entities;
    \item Examples of the most irrelevant candidates (used as negative prompts).
\end{itemize}

Based on this structured feedback, the agent adaptively revises the original prompt to construct a new, more focused prompt, allowing the LLM to better understand the semantics of the mention. Subsequently, the framework invokes the candidate generation module to reconstruct a new set of candidate. This process forms a closed loop of feedback–prompt revision–candidate retrieval, progressively improving the accuracy of the candidate set in each iteration. The framework sets a default maximum of 5 iterations. If the correct entity is not included after 5 iterations, the process terminates automatically and returns a \textit{matching-failed} signal.

\subsection{Entity-Clozer: Cloze-Style Entity Linking}
\label{sec:Entity-Clozer}

LLMs are deep neural network models trained on massive datasets, demonstrating superior capabilities in semantic understanding and reasoning tasks compared to traditional machine learning methods. To fully leverage their logical reasoning abilities, we reformulate the conventional string-matching-based entity linking task into an LLM-powered cloze-style reasoning task.

\subsubsection{Cloze-style Prompt}
% \textbf{Cloze-style Prompt. }
As illustrated in Figure~\ref{fig:overview}(IV), our framework employs the Entity-Clozer agent to construct a structured cloze-style prompt to guide LLMs in MEL. The prompt consists of four parts.
\begin{itemize}
    \item[\textbullet] [SENTENCE] represents a semantic summary of the target mention,
    \item[\textbullet] [TARGET MENTION] represents the mention to be linked,
    \item[\textbullet] [OPTIONS] provides all candidates from the candidate set $\mathcal{C}$,
    \item[\textbullet] [SELECT BEST OPTION] represents the requirements of the cloze task.
\end{itemize}

By processing this cloze-style prompt, the LLM performs semantic reasoning and selection over the candidates, enabling end-to-end MEL.

%%--------------------Experimental setting--------------------
\section{Experimental setting}
In this section, we conducted comprehensive experiments on five public MEL datasets to evaluate our proposed DeepMEL. Moreover, we present extensive analyses to provide a more profound understanding of our framework.

\subsection{Dataset}
Following previous works, we evaluated the performance of DeepMEL on five MEL datasets: WikiMEL, Richpedia, WikiDiverse, WikiPerson, and $M^3EL$. The following are the details of the datasets.

\begin{itemize}
\item \textbf{Wikidata-MEL}~\cite{ZhouWLXW21} contains over 18K multimodal samples extracted from Wikipedia\footnote{https://en.wikipedia.org} and Wikidata\footnote{https://www.wikidata.org}, the majority of the entity types in the dataset are Person.

\item \textbf{Richpedia-MEL}~\cite{ZhouWLXW21} is jointly constructed from Richpedia~\cite{Richpedia} and Wikipedia, containing over 17K multimodal samples.

\item \textbf{WikiDiverse}~\cite{wang2022wikidiverse} is a manually annotated MEL dataset consisting of diverse contextual topics and 13 entity types derived from Wikinews~\footnote{https://www.wikinews.org/}. 

\item \textbf{WIKIPerson}~\cite{WikiPerson} constructed based on Wikipedia for the Visual Named Entity Linking task. It contains approximately 16 million entities across 13 categories, and covers a wide range of topics.

\item \textbf{$M^3EL$}~\cite{WF-M3EL} is a large-scale and high-quality MEL dataset, manually labeled and constructed based on Kaggle\footnote{https://www.kaggle.com}, DBpedia\footnote{https://www.dbpedia.org}, Wikipedia, and Wikidata. It includes 79,625 instances, covering 9 diverse multimodal tasks and 5 distinct topics (e.g., Movie, General Knowledge, Person, Book, Sport).
\end{itemize}

We maintained the same dataset split consistent with previous work~\cite{UniMEL}. In both WikiMEL and Richpedia, the dataset is split into a 7:1:2 ratio for training, validation, and testing, respectively. We selected the test set for evaluation. In WikiDiverse, we selected the test set comprising 1,570 image-caption pairs. In WikiPerson, we selected the test set that contains 6,142 images, along with their labels (Wikidata Qid) and descriptions. In $M^3EL$, we also use the test set containing 7,962 instances with their labels (Wikidata Qid), context, images, and descriptions. The statistics of the datasets are summarized in Table~\ref{tab:Statistics_of_EX_dataset}.

\begin{table}[width=\linewidth,cols=4,pos=t]
\centering
\caption{Statistics of five widely used multimodal entity linking datasets, including Wikidata-MEL, Richpedia-MEL, WikiDiverse, WikiPerso, and $M^3EL$. These datasets are commonly used benchmarks for evaluating MEL tasks across diverse source domains and entity types.}
\label{tab:Statistics_of_EX_dataset}
\begin{tabular*}{\tblwidth}{@{} CCCC@{} }
\toprule
\textbf{Dataset} & \textbf{Source} & \textbf{Topic} & \textbf{Size} \\ \midrule
Wikidata-MEL & Wikidata & encyclopedia & 18k samples \\ \midrule
Richpedia-MEL & Richpedia & News & 17k samples \\ \midrule
WIKIDiverse & Wikipedia & News & 8k captions \\ \midrule
WIKIPerson & Wikipedia & Person & 50k images \\ \midrule
$M^3EL$ & \makecell{WikiPedia, Wikidata \\Dbpedia} & \makecell{Sports, Movies, Books, \\Person, Common} & \makecell{79K instances\\318.5K images} \\ \bottomrule
\end{tabular*}%
\end{table}

\subsection{Baselines}
We compared our method with recent state-of-the-art methods, which are divided into two groups: Textual-Only methods and Textual-Visual fusion methods. The following are the details of the baselines.

\begin{itemize}

\item \textbf{ARNN}~\cite{ARNN} (Textual-Only) captures textual features through a dual architecture comprising of Attention-RNN components.

\item \textbf{BERT}~\cite{BERT} (Textual-Only) encodes the mention context and entity description and then calculate the similarity score.

\item \textbf{BLINK}~\cite{wu2019scalable} (Textual-Only) uses a bi-encoder for candidate retrieval and a cross-encoder for candidate re-ranking.

\item \textbf{SumMC}~\cite{SumMC} (Textual-Only) uses zero-shot GPT-3 prompting for guided summarization and EL as a multiple-choice selection.

\item \textbf{LLMaEL}~\cite{LLMAEL} (Textual-Only) leverages LLMs as knowledgeable context augmenters, generating mention centered descriptions as additional input, while preserving traditional EL models for task specific processing..

\item \textbf{DZMNED}~\cite{SnapCaptionsKB} (Textual-Visual) employs a cross-modal attention mechanism to integrate visual, textual, and  features of character-level mentions.

\item \textbf{JMEL}~\cite{JMEL} (Textual-Visual) establishes modality alignment through fully connected layers that project both visual and textual features into a shared latent space.

\item \textbf{MEL-HI}~\cite{zhang2021attention} (Textual-Visual) implements a two-stage cross-modal filtering approach, where it first eliminates visually irrelevant images through image-text correlation analysis, then applies an attention mechanism to align mention-entity features in the remaining candidates.

\item \textbf{HieCoAtt}~\cite{HieCoAtt} (Textual-Visual) introduces a co-attention mechanism that constructs co-attention maps at multiple levels.

\item \textbf{ALIGN}~\cite{align} (Textual-Visual) features a dual-encoder architecture with EfficientNet as its vision encoder and BERT as its text encoder, and learns to align visual and text representations with contrastive learning.

\item \textbf{BLIP-2}~\cite{blip2} (Textual-Visual) leverages frozen pre-trained image encoders and large language models (LLMs) by training a lightweight, 12-layer Transformer encoder in between them, fusing visual, textual of mentions.

\item \textbf{InstructBLIP}~\cite{InstructBLIP}  (Textual-Visual) proposes a new vision-language instruction-tuning framework using BLIP-2 models, achieving good generalization performance on vision-language tasks.

\item \textbf{SigLIP}~\cite{siglip} (Textual-Visual) operates solely on image-text pairs and does not require a global view of the pairwise similarities for normalization.

\item \textbf{CLIP}~\cite{clip} (Textual-Visual) implements separate Transformer encoders for each modality, enabling joint learning of visual-textual feature correlations.
 %=============================
\item \textbf{GHMFC}~\cite{GHMFC} (Textual-Visual) employs gated hierarchical multimodal fusion and contrastive training to capture fine-grained cross-modal feature.

\item \textbf{MMEL}~\cite{MMEL} (Textual-Visual) learns the features of mention contexts and candidates at the same time by joint learning.

\item \textbf{DRIN}~\cite{XingZWLZD23} (Textual-Visual) models four types of alignment between a mention and an entity and constructs a dynamic GCN to select the appropriate relations for different input.

\item \textbf{DWE}~\cite{SongZ00LMW24} (Textual-Visual) enhances the target mention by refined multimodal information from Wikipedia.

\item \textbf{GEMEL}~\cite{GEMEL} (Textual-Visual) trains a feature mapper to enable cross-modality interactions.

\item \textbf{UiMEL}~\cite{UniMEL} (Textual-Visual) enhances the target mention and entities by textual and visual information.

\item \textbf{LlaVA}~\cite{Llava} (Textual-Visual) is an open-source chatbot trained by fine-tuning LlamA/Vicuna on GPT-generated multimodal instruction-following data.

\item \textbf{Qwen2.5-VL}~\cite{qwen2} (Textual-Visual) is a multimodal vision-language model with strong visual recognition capabilities, capable of identifying common objects, charts, layouts, and other elements.

\item \textbf{MiniGPT-4}~\cite{MiniGPT} is a large multimodal model based on deep learning, capable of processing multimodal data.

\item \textbf{GPT-4}~\cite{gpt4} (Textual-Visual) is a large multimodal model that can take image and text input and produce text output.

\item \textbf{DeepSeek}~\cite{DeepSeek} (Textual-Visual) performs very well on tasks such as mathematics, code, and natural language reasoning..

\end{itemize}

\subsection{Implementation Details}

\textbf{Knowledge Graph. } In our framework, we primarily consider the Wikidata, which is a knowledge graph complementing Wikipedia (providing rich encyclopedic information about world entities), which organizes and structures this knowledge in the form of triples, making the information more structured and actionable.

\textbf{Experimental Setups. } We use a series of LLMs, where the temperature parameter is set to 0.75 (for consistency in fixed output formats) and the maximum token length for the input is set to 512,  and other parameters remaining at their default settings. We use 32 shots in the semantic summary prompt and 2 shots in the cloze-style prompt and the maximum number of iterations was limited to 5 for the candidates adaptive iteration. Wikidata is used as the source KG, and the number of search results is set to 200.  

\textbf{Evaluation Metric. }We use Accuracy ( precision \textbf{@}1) to evaluate MEL effectiveness in all experiments.

\begin{equation}
    \text{Acc} = \frac{1}{N} \sum_{i=1}^{N} \mathbb{I}(M_i \leftrightarrow E)
\end{equation}

where $M_i$ is the target mention, $E$ is the correct entity, $N$ is the total instances, and the indicator function $\mathbb{I}(M_i \leftrightarrow E)$ is defined as:

\begin{equation}
    \mathbb{I}(M_i \leftrightarrow E) =
    \begin{cases}
    1, & \text{if } M_i \text{ matches } E \\
    0, & \text{otherwise}
    \end{cases}
\end{equation}

%%--------------------Results--------------------
\section{Results}
We continue to adopt some baseline results from~\cite{UniMEL} and use five publicly available datasets. The specific results of our proposed DeepMEL method against several competitive approaches are presented in Table~\ref{tab:Main-Results}.

\begin{sidewaystable}[tbp]
\renewcommand{\arraystretch}{1.25}
\caption{Performance comparison of different methods on five MEL datasets from 3-10 candidates. The best score is highlighted in bold.}
\label{tab:Main-Results}
\begin{tabular*}{\textheight}{@{\extracolsep\fill}c|c|c|c|c|c|c}
\toprule
\textbf{Modality} & \textbf{Model} & \textbf{Richpedia-MEL} & \textbf{Wikidata-MEL} & \textbf{WikiDiverse} & \textbf{WikiPerson} & \textbf{$M^3EL$} \\ \midrule
\multirow{5}{*}{\textbf{\makecell{Textual — Only}}} & BERT & 0.32 & 0.32 & 0.22 & - & 0.33 \\
 & BLINK & 0.31 & 0.31 & - & - & 0.34 \\
 & ARNN & 0.31 & 0.32 & 0.22 & - & 0.34 \\
 & SumMC & 0.44 & 0.41 & 0.36 & - & 0.46 \\ 
 & LLMaEL & 0.42 & 0.45 & 0.37 & - & 0.48 \\ \midrule
\multirow{20}{*}{\textbf{\makecell{Textual — Visual}}} & DZMNED & 0.30 & 0.31 & - & 0.26 & 0.33 \\
 & JMEL & 0.30 & 0.31 & 0.22 & 0.25 & 0.34 \\
 & MEL-HI & 0.35 & 0.39 & 0.27 & 0.25 & 0.38 \\
 & HieCoAtt & 0.37 & 0.41 & 0.28 & 0.27 & 0.41 \\
 & GHMFC & 0.39 & 0.44 & - & 0.29 & 0.42 \\
 & MMEL & - & 0.72 & - & - & 0.54 \\
 & ALIGN & 0.46 & 0.58 & 0.70 & 0.72 & 0.67 \\
 & BLIP-2 & 0.55 & 0.42 & 0.67 & 0.65 & 0.73 \\
 & SigLIP & 0.54 & 0.44 & 0.76 & 0.67 & 0.71 \\
 & CLIP & 0.52 & 0.36 & 0.42 & 0.55 & 0.74 \\
 & InstructBLIP & 0.62 & 0.38 & 0.47 & 0.54 & 0.73 \\
 & DRIN & - & 0.66 & - & - & 0.74 \\
 & DWE & 0.68 & 0.45 & 0.48 & 0.47 & 0.71 \\
 & DWE+ & 0.73 & 0.73 & 0.51 & 0.51 & 0.73 \\
 & GEMEL & 0.81 & 0.82 & 0.86 & 0.62 & 0.73 \\  
 & UiMEL & 0.84 & 0.83 & 0.81 & 0.64 & 0.75 \\ \cmidrule{2-7}
 & Qwen2.5-VL & 0.64 & 0.61 & 0.59 & 0.62 & 0.54 \\
 & LlaVA & 0.65 & 0.62 & 0.63 & 0.58 & 0.57 \\
 & MiniGPT & 0.66 & 0.61	& 0.55 & 0.59 &	0.62 \\
 & GPT-4 & 0.68 & 0.57 & 0.56 & 0.60 & 0.65 \\
 & DeepSeek & 0.70 & 0.63 & 0.64 & 0.66 & 0.69 \\ \cmidrule{2-7}
 & \textbf{DeepMEL} & \textbf{0.86} & \textbf{0.88} &\textbf{0.81} & \textbf{0.73}& \textbf{0.89} \\ 
 \bottomrule
\end{tabular*}
\end{sidewaystable}

\subsection{Comparison with Previous Textual-Only EL Models. }
DeepMEL outperforms other Textual-Only Entity Linking models across five datasets, demonstrating superior performance. In comparison, SumMC and LLMaEL also perform well, with accuracy improvements ranging from 10\% to 15\%. This advantage can be attributed to SumMC and LLMaEL's use of large language models (LLM) to compress document content into concise sentences directly related to mentions, which not only enhances the contextual information around mentions but also boosts their ability to tackle more complex problems. However, both SumMC and LLMaEL have yet to fully address the crucial role of semantic disambiguation, which remains an area for further improvement.

In existing EL models, tasks with missing or incorrect options are often neglected and not effectively handled. In contrast, DeepMEL compensates for this by searching online Wikipedia data to obtain the most up-to-date and accurate information, using these search results as candidate options for the model. This approach not only fills the gaps left by traditional Textual-Only EL models but also significantly enhances the model's performance in handling more complex tasks. The experimental results demonstrate that Textual-Only methods still show a significant performance gap compared to state-of-the-art Visual-Textual fusion approaches. This gap primarily stems from the inherent limitations of Textual-Only models in processing ambiguous references and low-quality mentions, as the lack of visual modality assistance restricts their ability to comprehend complex semantics.

\subsection{Comparison with Previous Textual-Visual MEL Models. }
Among the different Visual-Textual methods, each demonstrates varying results, with differences mainly reflecting the way visual representations are extracted and how they are integrated with textual information. Among the top five Visual-Textual methods listed in Table 2, the GHMFC method performs well due to its fine-grained handling of cross-modal information learning. This suggests that shallow-modal interactions and simple multimodal fusion strategies may not significantly improve the performance of the multimodal emotion reasoning (MEL) task. Additionally, the CLIP series models, pre-trained on large-scale image-text corpora, have also achieved significant performance improvements. Notably, the DWE+ method, which leverages semantic enhancement from Wikipedia and fully utilizes visual information, outperforms all other baseline methods in datasets, second only to our proposed DeepMEL method.

\subsection{Comparison with Multimodal Large Language Models. }
Table~\ref{tab:Main-Results} presents the performance of various large language models (LLMs) on the MEL task. Specifically, mentions in the document are labeled as [mention], and the context of the mention, along with a set of candidates, are provided as input. The task of the model is to select the candidate most relevant to the [mention] based on its understanding of the context. The experimental results show that DeepSeek performs the best on the MEL task, with an average accuracy (Acc) higher by 5\%-15\% compared to other models. This advantage may be due to input text length limitations, as longer contexts may introduce noise, which could affect the accurate matching of the mention and reduce the MEL accuracy.

\begin{table}[htbp]
\caption{Ablation study results on the key modules of DeepMEL, where \textbf{w/o} (without) indicates the absence of a specific module.}
\label{tab3:DeepMEL-Versions}
\begin{tabular*}{\textwidth}{@{\extracolsep\fill}lcccccc}
\toprule
Model & Richpedia-MEL & Wikidata-MEL & WIKIDiverse & WIKIPerson & $M^3EL$ \\ \midrule
DeepMEL & 0.86 & 0.88 & 0.81 & 0.73 & 0.89 \\ 
w/o Entity-Clozer & 0.79 & 0.84 & 0.67 & 0.69 & 0.72 \\
w/o Candidate-Adapter & 0.73 & 0.78 & 0.61 & 0.62 & 0.64 \\
\quad\quad w/o Candidate Adaptive & 0.75 & 0.82 & 0.63 & 0.65 & 0.69 \\
\quad\quad w/o Candidate Search & 0.74 & 0.80 & 0.61 & 0.64 & 0.66 \\
w/o Modal-Fuser & 0.70 & 0.72 & 0.57 & 0.59 & 0.63 \\
\quad\quad w/o Visual Generation & 0.73 & 0.76 & 0.58 & 0.60 & 0.62 \\
\quad\quad w/o Contextual Distill & 0.71 & 0.73 & 0.59 & 0.60 & 0.61 \\
w/o Role-Orchestrator & 0.68 & 0.61 & 0.55 & 0.58 & 0.56 \\ \bottomrule
\end{tabular*}
\end{table}

\begin{table}[!b]
\centering
\caption{Performance comparison of DeepMEL with different LLMs and LVMs as agents. Under the fixed architecture of the DeepMEL, we evaluate the impact of replacing the textual and visual agent with different LLMs and LVMs.}
\label{tab4:different-LLMs}
\begin{tabular*}{\textwidth}{@{\extracolsep\fill}ccccccc}
\toprule
Modality & LLMs & Richpedia-MEL & Wikidata-MEL & WIKIDiverse & WIKIPerson & $M^3EL$ \\ \midrule
\multirow{5}{*}{\makecell{Textual \\ Agent}} & DeepSeek & 0.86 & 0.88 & 0.81 & 0.73 & 0.89 \\
& GPT-4 & 0.82 & 0.86 & 0.77 & 0.70 & 0.83 \\
& GPT-3.5-turbo & 0.78 & 0.79 & 0.75 & 0.65 & 0.74 \\
& Qwen2-7B & 0.69 & 0.70 & 0.63 & 0.59 & 0.66 \\
& LLaMA2-13B & 0.64 & 0.62 & 0.61 & 0.59 & 0.67 \\ \midrule
\multirow{4}{*}{\makecell{Visual \\ Agent}} & GPT-4 & 0.84 & 0.86 & 0.80 & 0.72 & 0.88 \\
& MiniGPT & 0.83 & 0.85 & 0.78 & 0.71 & 0.87 \\
& LlaVA & 0.77 & 0.72 & 0.73 & 0.64 & 0.69 \\
& Qwen2.5-VL & 0.66 & 0.69 & 0.61 & 0.55 & 0.62 \\ \bottomrule
\end{tabular*}%
\end{table}

\begin{table}[t]
\caption{Results under different key hyperparameter settings. \textit{BLEU} and \textit{BERTScore} are used as similarity metrics. \textit{Temperature} controls the randomness of the model's output. \textit{Candidates} refer to the initial number of candidates. \textit{Iterations} denote the number of iterations in Candidate-Adapter.}
\label{tab5:DeepMEL-Simis}
\begin{tabular*}{\textwidth}{@{\extracolsep\fill}ccccccc}
\toprule
\textbf{Similarity} & Richpedia-MEL & Wikidata-MEL & WIKIDiverse & WIKIPerson & $M^3EL$ \\ \midrule
BLEU & 0.82 & 0.74 & 0.76 & 0.67 & 0.80 \\ 
BERTScore & 0.86 & 0.88 & 0.81 & 0.73 & 0.89 \\ \midrule
\midrule
\textbf{Temperature} & Richpedia-MEL & Wikidata-MEL & WIKIDiverse & WIKIPerson & $M^3EL$ \\ \midrule
0.25 & 0.38 & 0.41 & 0.39 & 0.30 & 0.47 \\ 
0.5 & 0.47 & 0.49 & 0.47 & 0.39 & 0.55 \\ 
0.75 & 0.82 & 0.88 & 0.81 & 0.73 & 0.89 \\ 
1.0 & 0.75 & 0.80 & 0.72 & 0.66 & 0.79 \\ \midrule
\midrule
\textbf{Candidates} & Richpedia-MEL & Wikidata-MEL & WIKIDiverse & WIKIPerson & $M^3EL$ \\ \midrule
50 & 0.59 & 0.63 & 0.56 & 0.49 & 0.64 \\ 
100 & 0.64 & 0.69 & 0.62 & 0.55 & 0.70 \\ 
200 & 0.82 & 0.88 & 0.81 & 0.73 & 0.89 \\ 
300 & 0.79 & 0.84 & 0.77 & 0.69 & 0.85 \\ \midrule
\midrule
\textbf{Iterations} & Richpedia-MEL & Wikidata-MEL & WIKIDiverse & WIKIPerson & $M^3EL$ \\ \midrule
1 & 0.35 & 0.35 & 0.32 & 0.29 & 0.34 \\ 
3 & 0.56 & 0.59 & 0.54 & 0.46 & 0.57 \\ 
5 & 0.86 & 0.88 & 0.81 & 0.73 & 0.89 \\ 
10 & 0.82 & 0.87 & 0.81 & 0.72 & 0.82 \\ 
\bottomrule
\end{tabular*}
\end{table}

\subsection{Comparison the Different Agent of DeepMEL. }
To further investigate the effectiveness of each agent in the framework, we conducted ablation experiments on the model and its variants across five datasets. Table~\ref{tab3:DeepMEL-Versions} shows the experimental results. 1) After disabling the cloze-style module, the framework considers the first entity from the embedding retrieval as the answer, resulting in a significant drop in accuracy, particularly on the Wikidiverse and $M^3EL$ datasets with complex entity categories. The results demonstrate that the cloze-style module effectively improves the performance of LLMs in a unified task format. 2) After disabling the Candidate-Adapter module, we matched the mention with the candidates (directly retrieved results) by calculating their semantic similarity, which led to a performance drop, highlighting the importance of the Candidate-Adapter. 3) After disabling Visual Generation, when directly utilizing text and images for MEL, the results show that the model's performance significantly decreases on Wiki and MEL datasets with complex entity types. 4) Building on the previous point, disabling Modal-Fuser and relying solely on the mention name, context, and images for EL leads to a decline in model performance. This indicates that key contextual information and the structured output from images play a crucial role in enhancing the mention and provide valuable cues for constructing subsequent candidates. 5) Disabling the Role-Orchestrator further degrades the model's performance, and the results further validate the importance of each agent within the framework.

%show that each agent has a significant impact, further validating the effectiveness of modules such as modality fusion and candidate quality enhancement.

\subsection{Comparison of Media Agent in DeepMEL Using Different LLMs and LVMs. }
In the Table~\ref{tab4:different-LLMs}, to evaluate the effectiveness of different foundation models serving as agents within the DeepMEL, we independently replaced the textual and visual agents while keeping the remaining components fixed. Specifically:
\begin{itemize}
    \item In the comparison of different LLMs as the textual agent, the visual agent was fixed as LLaVA.
    \item In the comparison of different LVMs as the visual agent, the textual agent was fixed as DeepSeek.
\end{itemize}

This experimental design ensures that the results accurately reflect the individual impact of each LLM or LVM on the overall performance of DeepMEL, without interference from confounding variables.

The \textit{Textual Agent} of Table~\ref{tab4:different-LLMs}, it can be observed that DeepSeek achieves the best performance in all datasets, with scores of 0.86, 0.88, and 0.89 in Richpedia-MEL, Wikidata-MEL and $M^3EL$, respectively—significantly outperforming other LLMs. GPT-4 ranks second, demonstrating stable performance across multiple datasets, such as 0.86 on Wikidata-MEL, slightly below DeepSeek. These results indicate that the quality of pre-trained large language models, the breadth of their training data, and their instruction-following capabilities have a direct impact on DeepMEL’s textual generation module. Models like DeepSeek and the GPT series exhibit strong alignment and contextual understanding, resulting in more accurate and context-aware outputs. In contrast, open-source models such as Qwen2 and LLaMA2, although large in parameter scale, lack sufficient instruction tuning or reasoning control, leading to significantly degraded performance on average, around 15\% – 20\% lower than the top-performing models.

The \textit{Visual Agent} of Table~\ref{tab4:different-LLMs}, GPT-4 achieves the highest scores in most datasets, for example, in $M^3EL$, it reaches a score of 0.88, which is 0.01 higher than MiniGPT, 0.19 higher than LLaVA, and 0.26 higher than Qwen2.5-VL, which demonstrating superior capabilities in visual-language alignment and cross-modal reasoning. MiniGPT ranks second, performing slightly better than LLaVA on several datasets; for instance, on WIKIDiverse, MiniGPT outperforms LLaVA by 0.05. In contrast, Qwen2.5-VL consistently achieves the lowest scores in all datasets, trailing GPT-4 by approximately 0.18 to 0.26, indicating limitations in visual grounding and entity understanding. Overall, the quality of multimodal pretraining, semantic alignment mechanisms, and the capacity to model entity attributes significantly impact DeepMEL's performance in MEL tasks. Notably, in scenarios involving complex structures or context-rich inputs, models with strong reasoning abilities and high-fidelity visual understanding play a critical role in boosting the overall framework's effectiveness.

\subsection{Comparison of Different Key Hyperparameter Settings.}
In Table~\ref{tab5:DeepMEL-Simis}, we use BLEU~\cite{bleu} and BERTScore~\cite{sentence-bert} to compute $sim(M_d, C_d)$ for filtering candidates, where $M_{d}$ represents the mention description after modal fusion, and $C_{d}$ is the attribute of the candidates in Wikidata. The experimental results show that using BERTScore significantly outperforms BLEU, with an average accuracy improvement of 15\%. Based on the experimental results from Table~\ref{tab3:DeepMEL-Versions} and Table~\ref{tab5:DeepMEL-Simis}, the quality of the candidate set is notably enhanced after semantic filtering. This method effectively addresses the shortcomings of small pre-trained language models in terms of candidate accuracy and coverage, which arise due to their relatively weaker semantic understanding capabilities.

From the \textit{Temperature} section of the Table~\ref{tab5:DeepMEL-Simis}, it is evident that setting the temperature to 0.75 yields the best overall performance for DeepMEL in all datasets. For example, on Richpedia-MEL and $M^3EL$, the scores reach 0.82 and 0.79, respectively, which represents significant improvements of +0.44 and +0.32 compared to the scores at temperature 0.25. Temperature controls the randomness of the model output, when the temperature is too low (e.g., 0.25), the model tends to produce conservative and repetitive outputs, often falling into rigid patterns that impair accurate entity selection. Conversely, overly high temperatures (e.g., 1.0) introduce excessive diversity, which can lead to semantic drift or hallucinations, resulting in incoherent or incorrect entity matching. Therefore, a moderate temperature setting (e.g., 0.75) achieves a desirable trade-off between determinism and diversity, substantially improving generation quality and alignment accuracy in the MEL task.

In the Candidates setting, increasing the initial number of candidates from 50 to 200 leads to a significant performance improvement. For example, the performance on Wikidata-MEL and $M^3EL$ improves by 21\% and 25\%, respectively, indicating that a larger candidate pool increases the likelihood of including the correct entity. However, when the number of candidates is increased further to 300, a slight performance drop is observed—for instance, an 8\% decrease on WIKIPerson and a 4\% decrease in $M^3EL$. This suggests that an excessive number of candidates may introduce noise or redundant information, resulting in longer texts and more ambiguous semantic spaces, which in turn hinders the model's ability to effectively model context. Therefore, setting an appropriate number of initial candidates helps balance entity coverage and model comprehension, making it a preferable trade-off under the current task setting.

%%--------------------Conclusion--------------------
\section{Conclusion}
In this paper, we propose a novel framework DeepMEL, which establishes a new paradigm for handling MEL tasks using a multi-agent approach. Specifically, we leverage the summarization capabilities of LLMs and the visual question\&answering capabilities of LVMs to enhance mention representations by fusing textual and visual information. We then employ tool-based search and adaptive strategies to dynamically filter the candidate set. Finally, we reformulate entity linking as a cloze-style task, utilizing the reasoning abilities of LLMs to select the most appropriate entity from the candidate set for a target mention. Extensive experiments conducted on five public datasets demonstrate the effectiveness of our DeepMEL compared to several state-of-the-art baselines.

\section*{CRediT authorship contribution statement} 
\textbf{F W }: Conceptualization, Methodology, Software, Writing - original draft. \textbf{T Y }: Conceptualization, Investigation - review \& editing. \textbf{Z Y }: Supervision, Investigation - review \& editing. \textbf{M H }: Supervision, Funding acquisition.
\textbf{J Z }: Supervision. \textbf{Z L }: Supervision. \textbf{X B }: Supervision.

\section*{Acknowledgments} This work was supported by the National Natural Science Foundation of China (No. 62476283, No. 62376284).

% \bibliographystyle{cas-model2-names}
\bibliographystyle{apalike}

\bibliography{cas-refs}

\end{document}